\documentclass[journal]{IEEEtran}

\usepackage[pdftex]{graphicx}
\DeclareGraphicsExtensions{.pdf,.jpeg,.png}
\usepackage{mathtools}

\ifCLASSINFOpdf
\else
\fi

\usepackage{amsmath}
\usepackage{xcolor}
\usepackage{algorithmic}
\usepackage{dirtytalk}
\usepackage{hyperref}

\usepackage{framed}
\usepackage{enumitem}
\hypersetup{colorlinks,allcolors=red}
\usepackage{setspace}
\usepackage{notoccite}
\usepackage[normalem]{ulem}

\usepackage{float}
\usepackage{graphicx,wrapfig,lipsum}
\usepackage{enumitem}
\def\BibTeX{{\rm B\kern-.05em{\sc i\kern-.025em b}\kern-.08em
    T\kern-.1667em\lower.7ex\hbox{E}\kern-.125emX}}

\newcommand{\eref}[1]{(\ref{#1})}

\usepackage{longtable}
\usepackage{cite}
\usepackage{array}
\usepackage{amsmath,amssymb,amsfonts}
\usepackage{algorithmic}
\usepackage{graphicx}
\usepackage{textcomp}
\newcommand{\virgolette}[1]{``#1''}
\definecolor{lime}{HTML}{A6CE39}
\usepackage{tikz,xcolor,hyperref}
\DeclareRobustCommand{\orcidicon}{%
	\begin{tikzpicture}
	\draw[lime, fill=lime] (0,0) 
	circle [radius=0.16] 
	node[white] {{\fontfamily{qag}\selectfont \tiny ID}};
	\draw[white, fill=white] (-0.0625,0.095) 
	circle [radius=0.007];
	\end{tikzpicture}
	\hspace{-2mm}
}

\foreach \x in {A, ..., Z}{%
	\expandafter\xdef\csname orcid\x\endcsname{\noexpand\href{https://orcid.org/\csname orcidauthor\x\endcsname}{\noexpand\orcidicon}}
}

\title{Learning Skills from Demonstrations: A Trend from Motion Primitives to Experience Abstraction}
\author{Mehrdad~Tavassoli\orcidA{},~\IEEEmembership{Student Member,~IEEE,}
        Sunny~Katyara\orcidB{},~\IEEEmembership{Student Member,~IEEE,}
        Maria~Pozzi\orcidC{},~\IEEEmembership{Member,~IEEE,}
        Nikhil~Deshpande\orcidD{},~\IEEEmembership{Member,~IEEE,}
        Darwin G.~Caldwell\orcidE{},~\IEEEmembership{Fellow,~IEEE,}
        Domenico~Prattichizzo\orcidF{},~\IEEEmembership{Fellow,~IEEE,}

\thanks{M. Tavassoli, N. Deshpande, D. G. Caldwell and D. Prattichizzo are with Department of Advanced Robotics at Italian Institute of Technology, Genoa, 16163, Italy. (e-mail: name.surname@iit.it) (\textit{Corresponding Author: Domenico Prattichizzo)}}
\thanks{S. Katyara is with Robotics and Automation team at Irish Manufacturing Research, Dublin, D24, Ireland. (e-mail: sunny.katyara@imr.ie).}
\thanks{M. Pozzi is with Department of Information Engineering and Mathematics at University of Siena, 53100, Italy. (e-mail: maria.pozzi@unisi.it).}
}

\begin{document}

\maketitle

\begin{abstract}
The uses of robots are changing from static environments in factories to encompass novel concepts such as Human-Robot Collaboration in unstructured settings. Pre-programming all the functionalities for robots becomes impractical, and hence, robots need to learn how to react to new events autonomously, just like humans. However, humans, unlike machines, are naturally skilled in responding to unexpected circumstances based on either experiences or observations. Hence, embedding such anthropoid behaviours into robots entails the development of neuro-cognitive models that emulate motor skills under a robot learning paradigm. Effective encoding of these skills is bound to the proper choice of tools and techniques. This paper studies different motion and behaviour learning methods ranging from Movement Primitives (MP) to Experience Abstraction (EA), applied to different robotic tasks. These methods are scrutinized and then experimentally benchmarked by reconstructing a standard pick-n-place task. Apart from providing a standard guideline for the selection of strategies and algorithms, this paper aims to draw a perspectives on their possible extensions and improvements.           
\end{abstract}


%
\IEEEpeerreviewmaketitle

\section{Introduction}

\IEEEPARstart{O}{ver} the past decade, the nature of tasks assigned to robots has changed, as has their workspace. Dynamic elements in their environment, like humans, objects, and other pieces of machinery, introduce various uncertainties to programming. Hence, to some extent, robots’ autonomy would be forfeited, obliging them to learn how to react to events using experiences or skills that have been transferred to them.

Traditionally, robots pre-program the necessary knowledge to execute their tasks. 
If the robot's behaviour is pre-planned, it might overfit the given task, substantially reducing the robot's self-sufficiency. Moreover, robots' environments have an abundance of dynamics, making it infeasible to hardcode every possible scenario. Instead, robots should learn the requisite skills to exchange with other elements around them. In this regard, learning algorithms evaluate every possible solution within the robot's reach. In such assessments, complexity grows exponentially for every extra Degree of Freedom (DoF) and is time-consuming. The computational dilemma mostly burdens robots with multiple DoFs like humanoids (e.g, WalkMan with 33, ASIMO with 34, and TALOS with 34, \cite{Schaal1999}).

\begin{figure}[t]
      \centering
      \includegraphics[width=8.5cm]{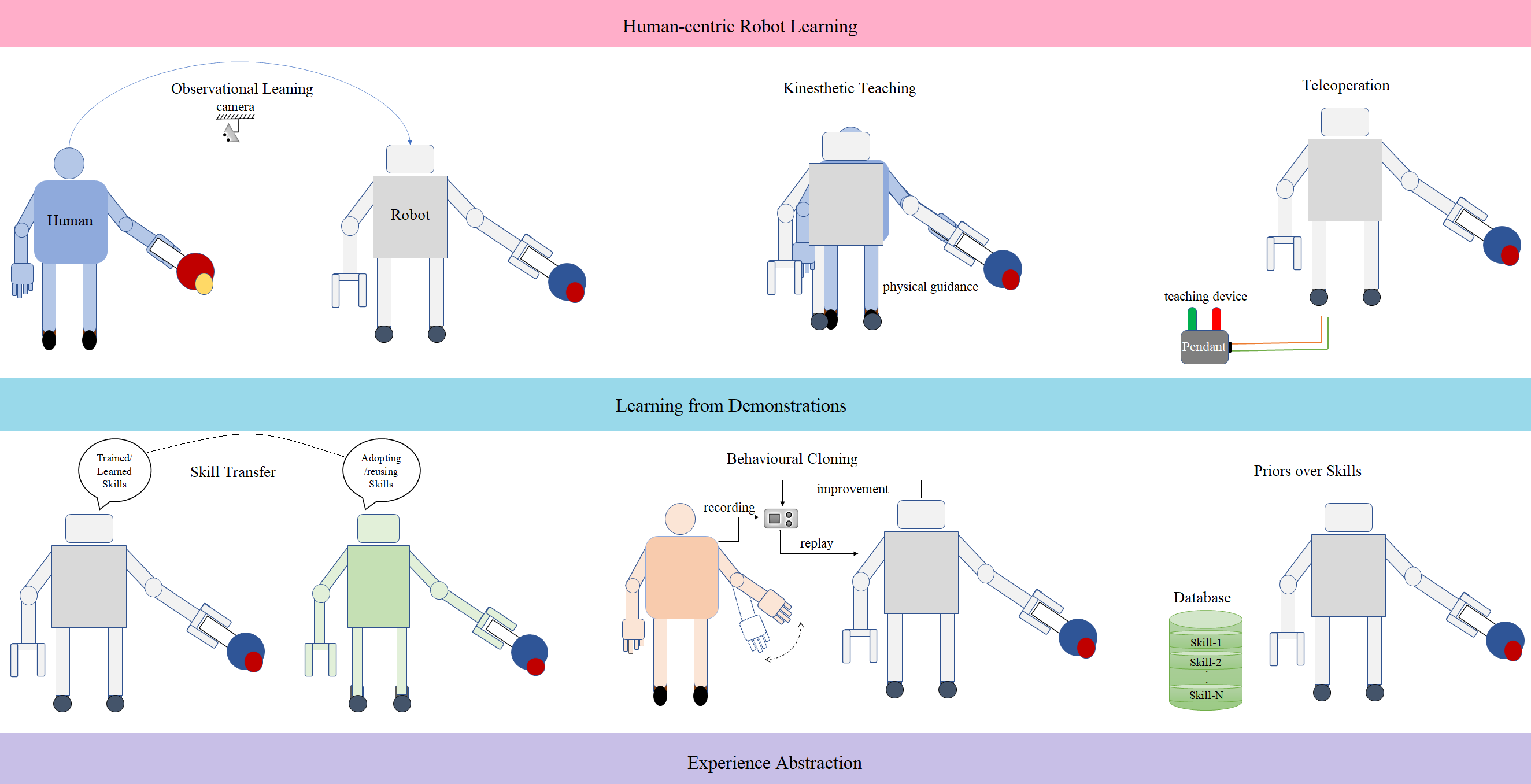}
      \caption{Different paradigms of human-centric robot learning. Observational Learning – Humans show how to perform a task, and robot vision infers it, Kinesthetic Teaching – Humans physically guide the robot towards task execution, Teleoperation – The robot is actuated using its teach pendant for task demonstration, Skill Transfer – A robot shares its experience with another robot for generic task realization, Behavioural Cloning – Humans initialize the robot’s policy with related actions of the task, Priors over Skills – Robots learn basic primitive skills from off-line data for generalized task}
      \label{onee}
      \vspace{-15pt}
   \end{figure}

Taking advantage of the previous approaches, a hybrid method called Behaviour Cloning\footnote{Behaviour cloning is known under different terminologies in the literature like Learning from Demonstration (LfD), Programming by Demonstration (PbD), and Imitation learning} (BC) was devised. It reduces computational complexity by shrinking the learning problem to a subset of the whole solution space allowing the operator demonstrates the task through mediums such as Kinesthetics, observations, and teleoperation. Afterwards, the robots incrementally learn elaborative movements based on prior knowledge (Bayesian Inference) \cite{Si2021}. Although shrinking the state-action search space nurtures training quantitively, optimal results are contingent on choosing the proper initial conditions \cite{Pastor2011}, as conceptually depicted in Fig. \ref{onee}.

Generally, Learning from Demonstration (LfD) can either encode the trajectory (low-level representation) or abstract the task by an array of sub-goals known as key points (high-level representation) \cite{Billard2016}. High-level representations are usually favoured due to their generality. However, they fail to scale the learning algorithm well to systems with multiple DoFs \cite{Schaal1999}. High-level representations were first introduced under \virgolette{Symbolic reasoning} terminology \cite{LozanoPerez1983}, which encapsulates many aspects of the task, including task sequences and their underlying hierarchy~\cite{Billard2016}. The proper action is chosen at each stage (state), leading to discrete state-action-state transitions \cite{Schaal1999}. Each state then turns into an \virgolette{if-then} statement to include the action's variability~\cite{Billard2016}. Nonetheless, for a general outcome, the inter-state transition should be continuous~\cite{Muench1994}, which leads to the concept of Movement Primitives (MP).

Movement Primitives (MP) are simple reusable units of movement where each unit can be the building block of a more complex movement. MP holds a complete state-action representation through a limited number of parameters, assisting the learning process~\cite{Schaal2005}. Hence, MP forms a modular control architecture. In many MP realizations, the learning algorithm scales up or down a specific MP block (predefined basic movement) to the segments of complex tasks. The composition of these blocks (in series or parallel) renders a complex task reproduction in new circumstances. MP methods can be categorized in multiple ways. Based on their tasks realization, MP approaches are divided into two main classes~\cite{Paraschos2018}: (i) state-based MP~\cite{Calinon2010,KhansariZadeh2011} and (ii) trajectory-based MP \cite{Nakanishi2004,Rozo2013,Rueckert2013}. In trajectory-based MP, temporal inputs are explicitly fed to generate the actions. In state-based MP methods, the temporal information and task sequences are either implicit (e.g. Hidden Markov Models (HMM)) or entirely ignored (e.g. in early versions of Gaussian Mixture Models (GMMs)).

\begin{figure}[t]
      \centering
      \includegraphics[width=8.5cm]{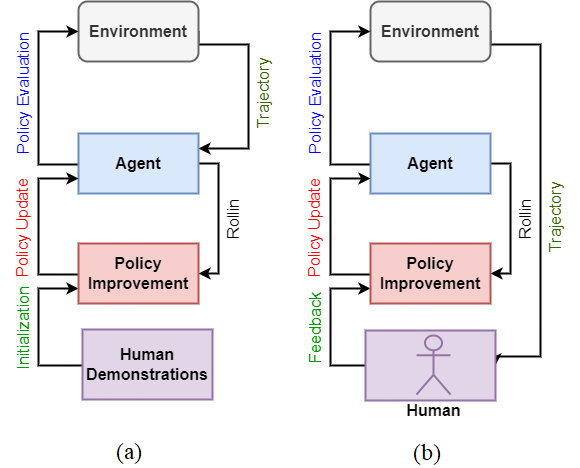}
      \caption{Block diagram representation of experience abstraction in imitation learning, (a) Offline cogitation of human skills from recorded demonstrations, (b) Online extraction of human behaviours through continuous interactions.}
      \label{EA}
      \vspace{-15pt}
   \end{figure}

MP approaches can be analyzed from different perspectives, namely the number of parameters and their applications. Some MP methods employ a finite number of parameters in their formulation, e.g. Gaussian Mixture Model, whereas others may use infinite parameters, e.g. Gaussian Processes (GP). Hence, if the dataset expands, the number of parameters is fixed in parametric approaches but will increase accordingly in non-parametric cases~\cite{Schulz2018}. Although their training datasets are bigger, parametric approaches offer better flexibility. Depending on the application, MP methods either deduce the human's intention to act accordingly (Intention learning) or emulate the observed behaviours in different plots (Imitation learning). The former is mainly utilized in HRC (Human-Robot Collaboration), whilst the latter is common where robot should have some degree of mission autonomy.

In contrast to MP, Experience Abstraction (EA) leverages human behaviours to extract the skills and rewards for goal accomplishment. Different training strategies deliver EA algorithms with the desired behaviours towards the goal, considering the resources, constraints, and setup. EA intelligent agents can learn offline from human-driven data (behavioural cloning, transfer learning, skill priors and inverse reinforcement learning). They can also continuously exchange feedback with the demonstrator to be trained online by (direct policy learning)~\cite{Abbeel2004,Ziebart2008,Finn2016,Cheng2018}. The block diagrams in Fig.~\ref{EA} summarize the differences between the two approaches.

Online learning has some flaws, including demanding a human-in-the-loop, which imposes extra cognitive burdens on the operator. It also needs metrics to evaluate the human decisions about the policy update, though it has yet to be matured. With offline training, there already exists data banks \cite{Dasari2019, Cabi2019} where meaningful task primitives and human behaviours can be extracted and transferred. These methods also generalize well to different task domains, making them robust against many intrinsic and extrinsic uncertainties ~\cite{Gupta2019,Siegel2020}.

This survey is not intended as a catalogue of MP and EA implementations in imitation learning and reinforcement learning. It focuses on providing a general guideline and outlook that suits for the audiences with different understandings of robot learning. It has the following goals:
\begin{itemize}
    \item Insights on MP and EA approaches and their potential use in robotic applications;
    \item Empirical evaluation of the most common learning algorithms based on different performance metrics;
    \item Time and space complexities of the highlighted approaches, their key characteristics, and use-cases;
    \item Identify research directions for MP and EA in imitation and reinforcement learnings.
\end{itemize}

Authors recommend referring to the provided references for
further details on the implementation of the cited approaches.

\section{Movement Primitives}

 \begin{figure}[t]
      \centering
      \includegraphics[width=8.5cm]{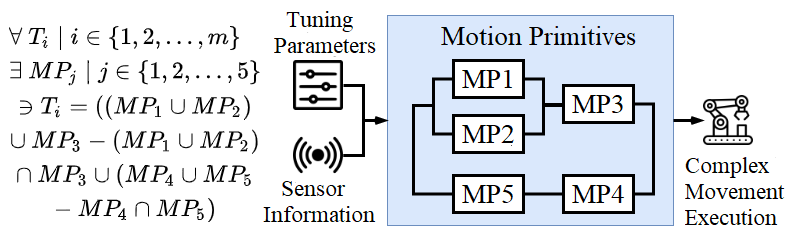}
      \caption{General representation of motion primitives as a sequence of series and parallel combinations of discretized movements for complex tasks.}
      \label{one}
      \vspace{-15pt}
   \end{figure}
   
Industrial robots mainly have three to six DoFs but in unstructured environments, they need as many DoFs as possible to generate a set of feasible solutions for their planning making their inference of the task and its constraints intractable~\cite{Asfour2008}. Hence, a modular structure with simple mechanisms of known behaviours is better. In turn, well-tuned parallel or series organization of these basic motor control units (motions) tailors the motion for the given task, reducing both time and computation. Known as Movement Primitives (MP), they were initially identified in animal studies~\cite{Shadmehr1994}, which pinpointed that the different regions in the neural circuit of a spinalized frog and their distinguishing functionality can be rearranged to form different movements.

For MP, primitive motion blocks $MP_j$ models an elaborate gesture or a complicated task $T_i$. As shown in Fig.~\ref{one}, tuning parameters and sensory information are viewed as inputs. Depending on the nature of the tasks to be performed, several formats of MPs have been proposed in the literature~\cite{Calinon2017,Huang2018,Kroemer2021}.


\subsection{Dynamic Movement Primitives}
Dynamic Movement Primitives (DMP) is a deterministic abstraction of MP that model each moving target as a simple spring-damper system. DMP-based approaches were originally introduced in~\cite{Schaal2006} and later modified in~\cite{Ijspeert2013}. DMP models each MP as a combination of a well-specified stable dynamical system coupled with a nonlinear modulation term known as forcing function. The forcing function tries to bring the actual system as close as possible to a desired imaginary trajectory called attractor field. Based on the task's specifications, there are two types of attractors~\cite{Ijspeert2002}:  pointwise attractors for discrete tasks, and limit-cycle attractors for periodic movements. The dynamical system in DMP  is described by \ref{eq:1}.

\begin{equation}\label{eq:1}
\begin{aligned}
    \tau \ddot{y}=\alpha_y(\beta_y(g-y)-\dot{y})+f \\ \tau \dot{z}=\alpha_z(\beta_z(g-y)-z)+f,
    \end{aligned}
    \footnote{Replacing y with z makes equation first order and more tractable }
\end{equation}
where $\alpha$ and $\beta$ are the constant gains, $g$ and $y$ are the system's target and current states, respectively.

By default, the forcing function $f$ is explicitly time-dependent, hampering the chance of coupling or coordinating with the remaining DoFs~\cite{Ijspeert2013}. A first-order dynamical system ($\dot{x}=-\alpha_xx$ for discrete motion and $\tau\dot{\phi}=1$ for rhythmic motion \cite{Ijspeert2013}), called Canonical system, implicitly encodes the temporal information to eliminate such time-dependency of $f$. A phase variable $x$, defined by \eref{eq:2} and \eref{eq:3}, respectively, for discrete and rhythmic movements, is assigned for this purpose. While $x = 1$ indicates the system's initial state, $x = 0$ marks the end of the task's execution. Furthermore, $x$ acts as an activation node that terminates each task when it reaches to its end, making the system stable. Typically, a single canonical system is shared through a limb (e.g., a humanoid's arm) to coordinate DMP for multiple DoFs~\cite{Ijspeert2013}.
 
\begin{equation}\label{eq:2}
    f(x)=\frac{\sum_{i=1}^N\Psi_iw_i}{\sum_{i=1}^N\Psi_i}x(g-y_0)
\end{equation}
\begin{equation}\label{eq:3}
    f(r,\phi)\footnote{Periodicity is included in the choice of canonical system but can be in the basis function as well \cite{Ijspeert2013}. For further understanding, please refer to~\cite{Schaal2006}. }=\frac{\sum_{i=1}^N\Psi_iw_i}{\sum_{i=1}^N\Psi_i}r
\end{equation}

Equations \eref{eq:2} and \eref{eq:3} are canonized forcing functions of pointwise and periodic (rhythmic) attractors in the same order.  $\Psi_i$ and $w_i$ are the $i^{th}$ basis function and weight $\forall i\in\{1,...,N\}\footnote{N is the total number of basis function}$. In \eref{eq:3}, $r$ is the amplitude, and $\phi\in [0,2\pi]$ is the phase angle for periodic oscillation.

As far as learning is concerned, different regression approaches like GMR~\cite{Calinon2007} or GP~\cite{Forte2012} can be applied. In~\cite{Schaal2006,Ijspeert2013}, a Locally Weighted Regression (LWR)~\cite{Atkeson1997} is used. The features that need to be learnt can be high-level parameters ($\tau,g,y_0,r$), that scale up or down the task execution without changing its essence, or the weights $w_i$.

DMP has many appealing features. Besides simplicity in its implementation, it is flexible to adapt to new targets owing to its extrapolation capacity and the coordination it brings to the final movement. However, as a deterministic approach, DMP fails to consider uncertainties in target adaptation. Hence, it only returns a subset of the optimal solution (mean value)~\cite{Paraschos2013}. Since DMP follows the attractor field rather than the robot's actual trajectory, it only considers two via-points, the endpoints, ignoring the rest~\cite{Zhou2019}. The state transition in DMP is prompt, which creates discontinuities in the velocity profile, leaving a jerk in its acceleration~\cite{Paraschos2018}. Lastly, concurrent activation of distinct dynamical systems is impossible, hindering the multi-tasking capability of DMP
~\cite{Pervez2018}.

To address these problems, DMP was modified by Hoffmann et al. ~\cite{Hoffmann2009} to give better modulation to new targets and become noise resilient. They derived an improved dynamical equation which is invariant to the choice of reference frame. The use of an acceleration term facilitates obstacle avoidance and reduces impulses compared to the initial version ~\cite{Schaal2006}. Apart from extending the use of DMP to include the transfer of skills from humans to the robot, Pastor et al. ~\cite{Pastor2009} also addressed the numerical issues faced by ~\cite{Hoffmann2009}. In addition, Stulp and Sigaud ~\cite{Stulp2012} minimized the noise effects on DMP's goal perturbation by subjecting the trajectory parameters ($\theta$) to noise and calculating the corresponding cost function. In the end, incremental learning with path integral $PI^{2}$ updates the parameters ($\theta$), which reduces the impact of noise on the trajectory. However, Meier and Schaal ~\cite{Meier2016} followed a different strategy reformulating the DMP to incorporate probabilistic implementation, encapsulating possible uncertainties.

\subsection{Probabilistic Movement Primitives}

Probabilistic Movement Primitives (ProMPs) are the probabilistic extension of DMP, which can adapt to varying starting and goal points. On top of that, ProMP modulates the via-points to avoid obstacles at a non-zero velocity profile. In its formulation, ProMP models the time-dependent variability of the trajectories using basis function representation. This representation is favoured over the simple distribution of Gaussians as it is easier to train and get a time-continuous method ~\cite{Paraschos2018}. Like DMP, ProMP utilizes Squared-Exponential (SE) basis functions representing discrete motions and Von Mieses basis functions for the rhythmic ones. Rather than driving just a subset of optimal solutions (mean value) in DMP ~\cite{Paraschos2018}, ProMP spans the whole training set for a range of solutions. The variability of the solution furnishes robust outcomes. At each time instant $t$, ProMP defines trajectory distribution based on the Hierarchical Bayesian Model (HBM). Its parameters determine the probability distribution of a specific trajectory stemming from its corresponding weights $p(\tau|w)$. The probability distribution of weights over variability parameters $\theta = \{\mu_{w}, \Sigma_{w}\}$ interactively captures the variance information ($p(w|\theta)$). At the end, marginalizing the weights out of trajectory distribution leads to the mentioned hierarchical architecture $p(\tau|\theta)=\int{p(\tau|w)}{\times}p(w|\theta){\times}d(w)$.

Like DMP ~\cite{Schaal2006}, ~\cite{Ijspeert2002}, ProMP also eliminates the explicit time dependency by a phase variable $z$. ProMP reconstructs the motion and enhances its robustness by considering the areas with high variability. Variability allows ProMP to yield optimal solutions within the demonstration range. Moreover, ProMP can trigger multiple primitives concurrently (by Gaussian product of MPs) to couple the respective joints. Thus, it is possible to coordinate the movement within and between a robot's limbs (by covariance matrix). Activation function $\alpha$ enables a relatively smooth transition between two adjacent MPs to preclude harmful jerks' that can be seen in DMP \cite{Saveriano2021}.

ProMP modulates the new via-points by conditioning the MPs at each time step to reproduce the trajectory. It ensures the resultant trajectory passes through the via-point $p(w|x^{*})$, where $x^{*}=[y^{*}_{t}, \Sigma^{*}_{t}]$ is the state where the robot should be at the given time $t$. $y^{*}_{t}$ and $\Sigma^{*}_{t}$ stand respectively for the position and required precision of via-points. The new states update the value of ($\mu_{w}$ and $\Sigma_{w}$) until the trajectory is assigned to the final via-point (the trajectory beyond the current via point is not optimized). This procedure iteratively continues prior to the final via-point being modulated. The optimal trajectory is synthesized in this state to meet all the via-points (the movement is completed). ProMP uses a single Gaussian in each iteration, like repetitive path decoding by GMR. Hence, it has a unitary prediction over the horizon, streamlining the GMM-GMR method\cite{Calinon2007}. Thus, it yields a substantially more accurate and general task recomposition compared to\cite{Calinon2007}.

Since ProMP uses a single Gaussian in each iteration, it is impossible to accommodate more than one via-point at a time\cite{Paraschos2018}. Like GMM, ProMP modulates the via-points with respect to the global frame. Purely stochastic in essence, ProMP is incapable of extrapolation \cite{Huang2019}, i.e., ProMP can not modulate the arbitrary points located out of the demonstration set\cite{Zhou2019}. Besides, the conclusive result is bounded to a more extensive demonstration than the other MP-based methods to cover most of the possible scenarios; otherwise, ProMP is prone to error\cite{Zhou2019}. Eventually, ProMP needs the area with high variability in case of a single demonstration, the corresponding variance is zero, making it deterministic like DMP. Since ProMP needs to calculate inverse of the covariance matrix, numerical errors may arise if the data is not sufficiently big. In this case, ProMP even under-performs DMP. Besides, ProMP again faces numerical error if the number of parameters to be learned is disproportionately big compared to the dataset size\cite{Huang2019}. It is worth mentioning that ProMP, either model-free or not, is not suitable for tasks that are loosely coupled in time \cite{Paraschos2018}.

\subsection{Parameterized Movement Primitives}

One of the common hurdles in developing LfD is the extraction of consistent features from the human demonstrations \cite{Calinon2014} \cite{Guenter2007}. In response, the movements are parameterized in terms of projected trajectory into local frames $\{j\}$. Choosing local frames over the global one $\{O\}$ provides a better generalization over tasks. Although task parametrization is not necessarily twisted with the local frame adaptation, it is usually described as such. In fact, state-based approaches like original versions of mixture models (Gaussian Mixture Model (GMM) and many variations of Hidden Markov Model (HMM)) are also parametrizing the task through a neutral angle. Basically, what is usually known as Task-Parametrized (TP) methods refers to the upgraded version of their predecessors. TP models assign local frames to the landmarks, such as the relevant obstacles to the robot itself in the robot's workspace\footnote{TP approaches are not limited to the frames, but in mainstream robotic applications, they are predominantly.} \cite{Calinon2014}. TP models provides the robot with a detailed narrative and widen its vision of its environment and task features. Accordingly, making it easier to extract the shared features from the dataset to adapt it efficiently to the task. To drive the task constraints (the common features), TP-based approaches are used to encode the desired reference trajectory from the demonstrations, as well as the precision and coordination prerequisite tracking under dynamic environments. However, TP model implementation is not simple and creates challenges in how and where to designate frames. This lack of a systematic strategy, leads to more empirical approaches based on common sense.

A single Gaussian cannot capture dataset for a task such as the movement of robot's end-effector. As a result, a mixture of Gaussian distributions may apply, where each distribution locally captures a part of the whole. Mixture model is an umbrella term which includes Gaussian Mixture Model (GMM), Hidden Markov Model (HMM), Semi-Hidden Markov Model (SHMM) and others that stem from the same concept. Indeed, HMM can be considered a conditioned GMM \cite{Calinon2017}.

GMM-GMR pair is based on Gaussian distribution, which makes it probabilistic. It extracts the consistent tasks features and reproduces them in novel situations. To highlight the importance of the duo, most robotic tasks have stochastic traits. As a mixture model, GMM forms a linear superposition of Gaussian density functions \cite{Pignat2019}, where the likelihood of the observation belonging to it is governed by: $p(X| \mu , \Sigma) = \sum_{k=1}^K\pi_k\mathcal{N}(X| \mu_k , \Sigma_k) $. Here, $X$ is the observations and $\mathcal{N(.)}$ is the Gaussian probability density function. $\{\pi_k, \mu_k, \Sigma_k \}_{k=1}^K$ are K-component GMM's hyperparameters that stand for the prior probability, mean values and covariance matrix of the $k^{th}$ components, respectively. Upon GMM delivering the model, Gaussian Mixture Regression (GMR) synthesizes the generalized movement. Given an arbitrary input as a new observation, GMR estimates the mean value $\hat{\mu}$ as generalized output (trajectory) and the task constraints and covariance matrix estimation $\hat{\Sigma}$\footnote{Covariance matrix sets the required accuracy by the task as well}. In GMM-GMR's formulation, the dataset consists of input and output $X=(X^{In}, X^{Out})$. GMM is brought by the joint probability distribution of input and output. However, GMR estimates the marginal probability distribution of output conditioned to new instance of input $\mathbb{E}(p(X^{Out}|X^{In}))$.

Based on Markov chains theory, a different realization, called Hidden Markov Model (HMM), was devised. First introduced in \cite{Rabiner1986}. It has been widely used for voice recognition \cite{Rabiner1989} ever since. In Bishop’s book, a detailed description of its formulation and the advancements of HMM are provided \cite{Bishop2006}.

GMM-GMR has many merits although it has problems for many scenarios. In its initial variants, GMM could not encapsulate the sequences of a task~\cite{Calinon2012}. In~\cite{Calinon2013}, time was explicitly formulated (timestamp) to get the temporal modulation of the new instances of the task. Moreover, particularly in incremental learning, the demonstration of tasks should possess a certain level of resemblance~\cite{Sylvain2009}. Otherwise, the reassessment of the new tasks parameters is not going to be straightforward~\cite{Huang2019}. Consequently, it deprives GMM-GMR of extrapolating arbitrary instances of the task. Lastly, in its basic format, one of the most notable flaws is its incapability of encoding the orientation~\cite{Calinon2020}. To overcome this, Zeestraten et al.~\cite{Zeestraten2017} reformulated GMM-GMR in the Riemannian space, where calculations are done locally in the planes, tangent to the manifold. Generally, the original GMM-GMR would be suitable for representing relevant features for continuous state-spaces. However, instituting the transition between consecutive states and the duration to stay at each is also of importance.  
In essence, HMM falls into the mixture models category. Unlike GMM,  HMM's states either proceed to the adjacent one or hold their ground. Such transitions between states are defined by their corresponding transition probabilities assigned to them. Since its development, HMM underwent modifications e.g., Semi-Hidden Markov Model (SHMM) \cite{Huang1989} proposed to include the duration, which determines the time in each state. For simplicity, this paper addresses only the original method. In summary, HMM tries to solve one of the following problems \cite{Rabiner1989}: 
\begin{itemize}
    \item Probability of having specific observations given the model;
    \item Estimation of the most likely state sequence, given a set of observations and the model;
    \item Determination of model's parameters, given state sequence and corresponding observations.
\end{itemize}
   For more information on other versions of HMM, refer to \cite{Bishop2006}.

HMM, thanks to especially its Left-Right architecture\footnote{There are other implementations, but LR implementation is the mainstream.}, intrinsically modulates time \cite{Rozo2013a} to capture the inherent sequence in a task. Like most stochastic methods, HMM is suitable for tasks described by random probability distributions, such as teleoperation, manipulation, and locomotion. Since the uncertainties are included during its state estimation, HMM is resilient to their perturbations. Additionally, it can stochastically capture Spatio-temporal variability \cite{Dillmann2004}. HMM is a double stochastic process, which implies that its underlying hidden state takes another stochastic process to be visible~\cite{Rabiner1989}. Khokar in \cite{Khokar2014} took advantage of this feature (i.e. hidden states) to develop a teleoperation setup, which helps novice operators, as the system interprets the user’s intention, to telemanipulate an object. The hidden state stands for the user's intention regarding the pre-grasp poses\footnote{The user's intention is not visible}. Upon recognition, the system picks the intended object and then maps the corresponding pre-grasp pose. 

Nevertheless, HMM has its drawbacks. The discrete transition from one state to another leaves discontinuity in its velocity profile, which leads to a jerk in its acceleration contour. Therefore, Monte Carlo sampling and averaging techniques \cite{Schmidts2011} are applied to smooth this out. Furthermore, HMM is not as good as GMR in task reconstruction \cite{Zeestraten2016}. In the end, like GMM, HMM cannot extrapolate every arbitrary input, limiting its strengths to the boundaries of the demonstration set.

\subsection{Via-points Movement Primitives}

DMP and ProMP both lacked functionalities, failing to deliver a generic solution for task reproduction. Asfour et al.~\cite{Zhou2019} introduced Via-points Movement Primitives (VMP) which built of previous work, but also addressed DMP’s ignoring of intermediary via points and ProMP’s lack of extrapolation. VMP has an elementary trajectory $h$ as its backbone and a shape modulation term $f$ that gives the prior probability $\omega$ for the modulation (during the learning stage).

Given the current values of parameter $\omega$, the probability density function of an arbitrary via-point defines whether the point belongs to the dataset or not. If the probability density of the new goal provided by $\omega$ exceeds the predefined threshold $\eta$, VMP modulates the via-point using the shape modulation term $f$. Otherwise, it extrapolates the via-point by translating the whole distribution into the new point's location. Afterwards, the shape modulation term alters the trajectory's shape even further to bring it closer to the target.

Although improved by \cite{Hoffmann2009,Dragan2015,Meier2016}, DMP does not explicitly consider the via-points properly mainly due to its philosophy, i.e., to track the actual trajectory and follow an attractor. Based on the results obtained by Zhou et al. \cite{Zhou2019}, in the reconstruction of the alphabetical letters, the high variance regions do not have any significance. Consequently, the ProMP is more inclined to be erroneous than the VMP. Further, ProMP training takes more resources as it needs to cover the entire robot's workspace owing to its lack of extrapolation \cite{Paraschos2018}. However, as mentioned before, VMP has no issues residing the points out of its training set.

\subsection{Kernelised Movement Primitives}

Kernelised Movement Primitives (KMP) was introduced in~\cite{Huang2019}. When the dimension of input/output is high, the total number of the basis function rises sharply, making it not a straightforward problem to train. Hence, kernel functions relieve the direct usage of the basis functions in such scenarios. KMP, in its formulation, included extrapolation capacity to compensate for the neglected observation during the demonstration. This extends the robot's dexterity beyond the training set. Besides, KMP reconstructs a smoother trajectory compared to ProMP \cite{Huang2019}. KMP suffers from high computational prices like many other of its peers. Moreover, tailoring and implementing KMP based on the task properties is problematic, restricting KMP from further developments. More elaboration on the computational complexities of the kernel function in machine learning is presented in \cite{CesaBianchi2015}. Driving KMP takes two essential elements: (i) A probabilistic reference trajectory and (ii) a parametric trajectory. 

From the training set, approaches like GMR \cite{Calinon2007} extract the probabilistic reference trajectory from a joint probability distribution. By assigning a weight vector $w$, the parametric trajectory can be obtained as a linear combination of the weighted basis function $\Theta^{T}\times w$, where $\Theta=Blkdiag[\phi_{1} ... \phi_{B}]$ is the matrix representation of basis functions. KMP minimizes the Kullback-Leibler (KL) Divergence loss to track the reference trajectory and meets the via-points. KL Divergence is well explained in \cite{Kullback1997,Kullback1951}. KMP drives parameter variability into an optimization problem \cite{Huang2019}. However, in ProMP \cite{Paraschos2018}, they are fulfilled by multiplication of known probability. 

In general, two types of KMP exist based on the reference of the frame: \sout{(i)} Local and \sout{(ii)} Global. In global KMP, before the definition of the kernel $k(^{.} , ^{.})$, KMP needs to be initialized. Afterwards, conditioning the parametric trajectory gives the probabilistic reference trajectory. Finally, KMP delivers its prediction of $\mu, \Sigma, K, K^{*}$, where they stand for mean values, covariance matrix, kernel matrix ($K=[k(s_{i},s_{j})] \forall i,j \in \{1,2, ... ,N\})$ and kernel matrix corresponding to the enquiry point $S^{*}$ $(K^{*}=[k(s^{*},s_{i})] \ \ \forall i \in \{1,2, ... ,N\}$) in the same order. These values solve the two optimization sub-problems for $\mu$ $\&$ $\Sigma$ separately.

In Local KMP, some reference frames are locally assigned considering the task features. Demonstrations and operations are all projected into the local frames to localize the task instances of the robot to increase its local movement range \cite{Huang2019}. As discussed before, localization of the frames enhances the extrapolation span. Huang et al. in \cite{Huang2019} utilized the local version to augment dexterity in their setup.

Local KMP, like its standard version, can modulate a series of via-points. For modulation of a new query, points $m$ $\in$ $R^{M \times 1}$; local KMP starts with extending the reference trajectory to include new query points $D\in R^{N \times 1}$. Eventually, each via-point is modulated similar to a standard KMP. On top of all, KMP can blend different solutions (trajectories) for a task using the product of their reference trajectory distributions. Such superposition of trajectories brings about a weighted solution (weighted mean values and covariance matrix).

\subsection{Functional Movement Primitives}

Functional Movement Primitives (FMP) sets a range of correspondences instead of a single time-dependent trajectory over the data points of the demonstrations. FMP develops a model using first and second-order movements of the demonstrations to make it linear. The linear model recursively predicts the unobserved quantities from the known values.

Gaussian Processes (GP) is one of the popular approaches in this domain. GP was first mentioned in \cite{Williams1995} as a flexible non-parametric method. Owing to its non-parametric attribute\footnote{In non-parametric models, the number of parameters increases with the size of data \cite{Wang2020}}, GP has fewer hyperparameters to learn in its kernel function. To some extent, GP can be perceived as an extension of Gaussian Probability distribution (GPD). Its main applications are regression and classification problems; this paper focuses only on the regression ones. The authors suggest referring to~\cite{Hensman2015} for more details about classification applications.

GP conducts regression by defining a distribution over infinite possible functions that fit the given dataset. In other words, rather than random variables (Scalar or vectorial), FMP defines the distribution of the functions. In contrast to GP, a conventional non-linear regression method renders only a subset (mean value) of all feasible solutions. After training for each input in the training set, Given the priors, Gaussian Process Regression (GPR) predicts (posteriors) to execute regression. The priors and their respective evaluations are ingrained using kernels (also known as covariance functions). The algorithm iteratively updates the priors by substituting them with the most recent posteriors until parameters converge to their optimal values \cite{Wang2020}.

Being Kernel-based is an essential feature of GP, enabling it to smoothen out the trajectories. Being kernel-based also allows to alleviate the problem of extracting the proper underlying function in the regression tasks. However, GP as  KMP, has the issue of dimensionality. This problem arises when the dimensionality of features is proportionally bigger than the data size. Besides its dimensionality problems, picking the right kernel is complicated yet crucial. Depending on the task specifications, a different kernel function should be used to guarantee performance. The choice of kernels goes from a simple dot product ($|A|\cdot|B|\cdot cos(\theta)$) to harmonical ones. As far as GP is concerned, Squared-Exponential (SE) kernels, defined by \eref{eqt:5}, are the default choice \cite{Duvenaud2014} thanks to the universal property of exponential function \cite{Wang2020} and its convexity.

\begin{equation}
\label{eqt:5}
k(x_i,x_j)=\sigma_f^2{\times}exp(-\frac{(x_i-x_j)^T(x_i-x_j)}{2l})
\end{equation}

$\sigma$ and $l$ are hyperparameters of the SE kernel that not only scale up or down the function vertically and horizontally but also define the amount of data that can be presented in their corresponding sense. More information about the selection and tuning of hyperparameters has been provided in \cite{Duvenaud2014}.

As mentioned earlier, GP defines a distribution over function $f$, mapping the input space into some virtual space in the real domain $\mathfrak{R}$. $f: \mathfrak{X} \rightarrow \mathfrak{R}$. Thus, \eref{eqt:6} governs the marginal distribution over the function:

\begin{equation}
\label{eqt:6}
 \forall \ x_i\in\mathfrak{X} \ \exists \ p(f(x_1), ... , f(x_n))= \mathcal{N}(m(x),K(x)) 
\end{equation}

where $m$ and $K$ are mean function and covariance (Kernel) function respectively, defined by \eref{eqt:7}.

\begin{equation*}
\forall i,j\ {\in} \{1, ... , n \} \ \exists \ \Sigma=\mathbb{E}((f(x_i)-m(x_i))(f(x_j)-m(x_j)) 
\end{equation*}
\begin{equation}
\label{eqt:7}
={K(x_i,x_j) }\in \mathbb{R}^{n \times n} 
\end{equation} 
\begin{equation*}
\forall i\ \in \{1, ... , n \} \ \exists \    m(x)=\mathbb{E}(f(\bar{x}))=\ \mu (x_{i})\}\mu_\in\mathcal{R}^{n \times 1}
\end{equation*} 

The two functions are the building blocks of SE kernels $f(x)~ \sim \mathcal{GP}(m,K)$. In the presence of i.i.d noise $\varepsilon_i \sim \mathcal{N}(.|0,\sigma^2)$, the outcome would be:  $ \forall i\in\{1, ... , n\} \exists \  \mathcal{D}=\{ (X_i,f(X_i)+\varepsilon_i)\}$ GP predicts for a new goal, given a new arbitrary input and the dataset $\mathcal{D}$ \cite{Ghahramani2011}, defined by \eref{eqt:8}:

\begin{equation}
\label{eqt:8}
    p(y_{new}|x_{new},D)=\int(y_{new}|x_{new},f,D)p(f|D)df
\end{equation}

The predictive distribution is then determined using \eref{eqt:9}:

\begin{equation}
\label{eqt:9}
    p(y_{new}|x_{new},D) \sim \mathcal{N}(\mu_{new},\sigma^{2}_{new})
\end{equation}

GP is a flexible approach to modeling processes using kernels. However, its dexterity entails a high computational cost that is common amongst most-if-not all kernel-based approaches like KMP \cite{Huang2019}, as described in \eref{eqt:10}.

\begin{equation}
\label{eqt:10}
    \begin{aligned}
    \mu_{new} = K_{new,n}(K_{n}+\sigma^2I)^{-1}y \\
    \sigma^2_{new} = K_{new,new}-K_{new,n}(K_{n}+\sigma^2I)^{-1}K_{n,new}\sigma^2
    \end{aligned}
\end{equation}

The inversion of $(K_{n}+\sigma^2I)^{-1}$ term, in predictive distribution, is very expensive $\mathcal{O}(n^3)$ \cite{Snelson2007}. The higher level of complexity in GP is problematic if the data size rises. Apart from that, high computational cost requires a high demand for memory in the system $\mathcal{O}(n^2)$ \cite{QuinoneroCandela2005}. The other issue with GP, especially in its initial formulation, is its mostly stationary covariance function \cite{Rasmussen2001}. This usually happens when there is a discontinuity or any changes in variance throughout the function. Finally, the accuracy of the approach highly depends upon the right choice of kernel function type, which makes it not so easy to implement. Hence, designing a proper model for a process is longer and harder. To lift off the computational load, sparse approximations are used. Based on the scope of action, they usually fall into one of the three classes: Global GP sparse approximation (GGP), Local GP sparse approximation (LGP), and Hybrid sparse approximation \cite{Snelson2007}.

In Global approximation, a limited number of points called support points are selected that represent the observations in their vicinity well. Consequently, the model is created by only a fraction of the whole dataset $m<<n$. A subset of the training set is substituted with $m$ pseudo-inputs whose locations are assigned using gradient-based approximation. This approach reduces the complexity level to $\mathcal{O}(n^2)$ and $\mathcal{O}(n)$, respectively, for training and prediction \cite{Snelson2005}. Although the global approximation is swift in reducing the computational complexity, it cannot correctly approximate the processes with fast dynamics (when the trajectory is too wiggly). Moreover, the assignment and allocation of the support points are not straightforward in global sparse approximation \cite{Schneider2010}. Local versions of the spars approximation offer more robustness in the abstraction of the process and cover the mentioned weakness of the Global one.

In LGP spars approximation, the data set is divided into smaller regions. The gating network assigns a single GP known as an expert to its respective subspace of the whole task. Hence, the expert's contribution is limited to its surroundings. Eventually, the entire process is summarised by the superposition of all these experts along the path. Due to its composition is known as Mixture of Experts (ME) \cite{Jacobs1991}. LGP uses the point in its surroundings, making prediction easier \cite{Snelson2007}. In \cite{Tresp2000}, the authors assign $3m$ \footnote{m is the number of experts} GPs, where one is responsible for the experts and the other two are used to model the noises and a separate gating network. Since the whole dataset is used for the computations, the overall complexity would be $\mathcal{O}(n^3)$, which is even more computationally expensive. On the other hand, Rasmussen and  Ghahramani \cite{Rasmussen2001} divided the whole trajectory into $m$ equal regions, then assigned an expert for each. As a result, the complexity level dropped to $log({\mathcal{O}(n^3)})$. 

To bypass the flaws of assigning experts for each task, they have introduced the concept of infinite experts accompanied by a gating network based on the Dirichlet process. Nguyen et. al in \cite{NguyenTuong2008} examined the accuracy and efficiency of LGPs regarding their computational time. Afterwards, they validated their method on a real-time model-based controller of a fast-compliant robot. In short, LGP is applied to more extensive sample-sized data with higher variation around its adjacent datapoints. Apart from that, LGP is a suggested approach for online learning \cite{NguyenTuong2008}.

Weighting the pros and cons of both variants of sparsing GP against each other proves that a hybrid solution would be the \sout{suite} best. Taking advantage of both the domains, a hybrid solution is proposed by \cite{Snelson2007}. It addresses the non-static abstraction challenge that the GP faces. In their implementation of the heteroscedastic GP, the noise rate is proportional to the input during regression. Based on this approach, whilst a GP is responsible for capturing the noise-free process, the other GP is designated for the noise variance \cite{Kersting2007}. Hence, it can be concluded that, to some extent, Pseudo-input can make the performance of the GP better for non-static systems \cite{Snelson2005}.

\section{Experience Abstraction}

Experience abstraction (EA) incrementally encodes new task-related behaviours using prior knowledge to speed up the learning process. It is either about agents learning from their own experiences or exploiting others' behaviour to accommodate learning \cite{Zhou2020} efficiently over time. Besides getting overfitted to the task, Self-exploration takes longer to train. Hence, the actual task variables initialize the agent instead of random values. However, a proper balance between initialization and iterative encoding needs to be made to achieve the desired behaviour. In general, the agent can be either a value-based or policy-based algorithm. Agents would train by either evaluating the values of state-action pair $Q(S, A)$ or optimal policy {$\pi^*$} to select the effective actions \cite{Sutton2018}. Nonetheless, both are trying to solve a similar problem formulated by Markov Decision Process (MDP). An MDP for incremental learning consists of a tuple with five elements \{States (S), Actions (A), Rewards (R), Transitions ($\rho$), Discounts ($\gamma$)\}.

According to MDP, the agent tries to maximize the cumulative discounted reward to perform the given task optimally. To obtain greater discounted returns, the value-function/policy-network needs to be appropriately configured and initialized. The three well-known techniques to achieve it are; Transfer learning, Behavioural cloning and Priors-over-skills \cite{Levine2020}.

\subsection{Skill Transfer}

Transfer learning is an approach where a pre-trained agent is either extended to learn a new task by using existing model parameters or shares its trained configuration with other inexperienced agents for continual learning. 

Considering the sampling inefficiency of learning in complex environments, the learnt skills for one task are often transferred to others using different mechanisms, which improves learning efficacy. One common option concerning transfer skills would be initializing new policy parameters with learnt skills. Subsequently, the hyperparameters would be tuned to adjust the learnt task to the new task \cite{Pastor2009}. However, its application would be limited to scenarios where there are resemblances between the original and the new tasks. Otherwise, a subset of skills that retain semantic information is exploited \cite{Platt2019}. This type of transfer learning is substantially used by value-based agents, where the abstraction of state-space ignores irrelevant observations to task skills (distractors) and then generalizes to state-spaces of different sizes. In stochastic environments and continuous complex state-spaces, deep neural networks are used to abstract the given value function within the simulated environment. In this regard, domain randomization is used and adapted to a physical system within the bounded set of policy parameters \cite{Bousmalis2018}. Besides, the neural networks prioritize tracking the history of the state, action, and goal tuples over the state-action pairs. In this case, the latter only applies if it helps leverage the policy parameters across multiple goals \cite{Schaul2015}.

However, instead of transferring skills in a single instance, it is sometimes helpful to consider the problem as sequential learning under multiple instances. For instance, when the pre-trained neural network is used for different tasks, its weights will be frozen during training. After that, a few lateral output layers will be added per task requirements, which speeds up the learning process, reduces the computational burden, and improves the inference performance as well \cite{Rusu2016}. In addition, sequential learning is applicable where it is more expensive to explore than exploit for complex tasks. Therefore, it starts initially with a sparse representation of tasks. In an incremental learning paradigm, the agent gradually builds intuition using smooth gradient steps with positive rewards \cite{Svetlik2017}.

The commonly used algorithms based on this idea are:
 \begin{itemize}
     \item Guided Policy Search \cite{Levine2015}: deriving interpolation controllers for task instances and using them for generating data to train a policy network;
     \item Hindsight Experience Replay (HER) \cite{Andrychowicz2017}: fake play between the agent-state and hindsight-observation for a given task to end being in goal;
     \item Scheduled Auxiliary Control \cite{Riedmiller2018}: choosing from the hierarchy of pre-defined task actions and learning to reach a defined goal on-fly.
 \end{itemize}

\subsection{Behavioural Cloning}

In response to the exploration-exploitation dilemma of incremental learning, Behavioural Cloning (BC) was introduced. BC initializes the robot's policy, so it takes greedy actions from the state space at early iterations. In this regard, BC falls into the supervised learning paradigm to learn policy parameters, either as a state-to-action mapping $\pi: S \rightarrow A$ for deterministic or a state-action pair mapping of probabilities over actions: $\pi: S \times A \rightarrow R$ stochastic policies. Besides allowing the agent to mimic the behaviour of the demonstrator in novel conditions, this approach helps to infer the reward function that the agent is trying to optimize \cite{Abbeel2010} for inverse reinforcement learning.

Several studies have shown BC's efficacy in many robotic tasks while initializing incremental learning policies \cite{Ross2011} \cite{Torabi2018} \cite{Duan2017}. The earlier research in \cite{Schaal1996} \cite{Atkeson1997a} focused on using BC to understand model dynamics and reward functions from human demonstrations, which fosters learning optimal policies with better convergence and sample efficiency. Meanwhile, Kim et al. in \cite{Kim2013} applied BC during the policy iteration phase for value function approximation using constrained convex optimization. On the other hand, in \cite{Rana2017,Kormushev2011,Niekum2015}, BC merged with motion planning using probabilistic inference. Consequently, BC was furnished to learn, detect and recover from conflicting conditions. Moreover, it employed perceptual modalities such as vision and tactile sensors to scale the learnt behaviours in complex scenarios.

In this context, \cite{Hester2018}, and \cite{Nair2018} have introduced new algorithms called Deep Q-learning from Demonstrations (DQfD) and Deep Deterministic Policy Gradient from Demonstration (DDPGfD). These approaches sought to learn policies for Atari games (DQfD) and object manipulation tasks (DDPGfD). The n-step Q-learning loss function was applied to fulfil the Bellman Equation's optimality. For gradient-free agents, however, the Policy Optimization from Demonstrations (POfD) algorithm \cite{Kang2018} included expert skills. Accordingly, POfD minimizes the gap between the demonstrated and learned policies with adversarial learning objectives and sparse reward signals.
The Demonstration Augmented in Policy Gradient (DAPG) approach \cite{Lee2019} used human demonstrations for gradient-based agents. Demonstration deduces a BC loss as a pre-training step. Thus, applying weighted heuristic optimization bears an extended cost function. This cost function interpolates BC and policy gradient losses to train an optimal policy using the Natural Policy Gradient technique \cite{Grondman2012}.

\subsection{Priors over Skills}

Encoding priors over skills is a way to extract behaviours and skills from data without any reward or task information. These data are generated either by humans or agents interacting with their surroundings. They are used to solve the downstream tasks, where the action space of trained policy represents the set of extracted skills. With an increasing number of such skills, the time-space complexity of action rises. The resulting complexities compromise learning performance due to the need for enormous extrapolation within the space of derived skills \cite{Shankar2019}. This technique also gave rise to promising results for offline incremental learning settings, where the policy is deduced from the previous experience of the agent. Eventually, it was exploited to accelerate online incremental learning \cite{Pertsch2020}. However, the downside of this approach is the need to annotate the collected experience for inferring and leveraging the desired skills across multiple tasks.

Considering the time-space complexity of such intertask skill transfer methods, the stochastic latent variables are used to define priors over the skills. Thus, the need for extended policy blocks for representing more skills would be alleviated \cite{Hausman2018}\cite{Lynch2020}. These types of skill embedding facilitate a faster exploration of intelligent agents within the complex skill space. Additionally, they are applied to avoid overestimating action values Q(s, a), which eventually accelerates incremental learning for downstream tasks. These intelligent agents represent asymmetric masking of behaviours across the library of skills that enables generalization within encoded priors over skills \cite{Lee2018}. Besides, this approach helps to overcome the issue of biases of an intelligent agent towards the learnt behaviours and exploit skills for lifelong learning \cite{Fujimoto2019}.  

\section{Discussion}

This paper studies some of the most important movement primitives, and experience abstraction approaches and use the standard peg-in-hole task as a benchmark to evaluate their performances. The metrics detailed in the following are exploited to evaluate each method.

\begin{figure*}[t]
      \centering
      \includegraphics[width=18cm]{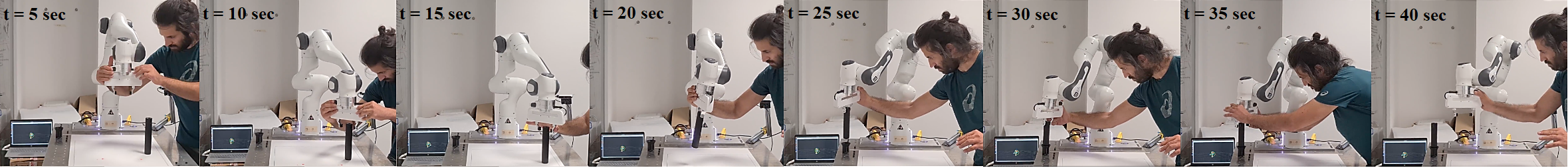}
      \caption{Human subject guiding physically a redundant robot arm to execute a peg-in-hole task. The corresponding 3D Cartesian space positional data is collected for training and evaluating different discussed models.}
      \label{demonstrator}
      \vspace{-15pt}
   \end{figure*}

\subsection{Evaluation Metrics}
\subsubsection{Time-Space Complexity}

Time-Space complexity is a paradigm for investigating the computational performance of an algorithm. It is governed by the time taken and memory consumed by an algorithm to execute the given set of instructions. The time and space constraints of an algorithm are competing interests; hence, a trade-off should be made between its efficiency (i.e., quicker and correct outputs) and finiteness (i.e., terminate and avoid memory leaks~\cite{Woeginger2004}). The asymptotic notations are used to compare and evaluate the time-space complexity of multiple algorithms, especially the Big-O notation ($\mathcal {O} $). It measures the change in the performance of two algorithms by increasing or decreasing the input sequence size. It defines the worst-case scenario of an algorithm. Mathematically, it is used to determine an upper bound of an algorithm. For instance, if an algorithm is defined by a function $f(N) = \mathcal{O}(g(N))$ and for some positive constants K and M, then the function can be written as $f(N) \leq K\times{g(N)}$ $\forall$ $N \geq M$. In this case, we are interested in how $f(N)$ increases as the size of the input $(N)$ increases.

\subsubsection{Root Mean Squared Error}

Root Mean Squared Error (rMSE) is a simple yet effective metric to evaluate the performance of prediction models. It measures an average Euclidean difference between the predicted and target values. rMSE penalizes even the small differences by squaring them, which leads to overestimating the trained model performance. Moreover, it is less prone to outliers~\cite{Willmott2005}. Mathematically, it is defined as $rMSE = \sqrt{\frac{1}{N}\sum_{t=1}^{N}({Y_{pred}}_t-{Y_{tgt}}_t)}$, where $N$ is number of samples, $Y$ is model output and $t$ represents current instance.

\subsubsection{Coefficient of Determination}

The coefficient of determination ($R^2$) defines the variation in predicted values compared to targets. The closer $R$ gets to $1$, the closer it captures $100\%$ of variance in the targets. To overcome false indications, specially over-fitting training data, the $R^2$ is adjusted using a number of independent variables, which then represent actual improvement \cite{Ozer1985}. Mathematically, it is expressed as $R^2=1-(\frac{N-1}{N-K-1})(\frac{\sum_{i=1}^{N}{(Y_{act}-Y_{pred})^2}}{\sum_{i=1}^{N}{(Y_{act}-Y_{mean}})^2})$, where N is total number of samples, K represents the number of independent variables, and Y is the output with respective actual, predicted and mean values.

\subsubsection{Information Loss}

Information theory measures both the amount of information present in a given dataset and that lost by the model after training. For example, the cross-entropy loss is commonly used in classification problems to evaluate the difference between the predicted and actual values \cite{Bishop2006}. Such losses can also be directly inferred by only considering the rules of information theory. In information theory, the measure of information loss is related to determining the entropy of a trained model or proposed algorithm \cite{Baez2011}. Mathematically, it is denoted as $H(X)=-\mathbb{E}[\log{p(X)}]$, where $p(X)=f_1(X)f_2(X)….f_n(X)$ is the probability distribution of random variable $X$ which is the output of the trained model. Each function component $f_i(X)$ is independent and contributes asymmetrically to the total information gained from $p(X)$. The greater the entropy of model $H(X)$, the greater the information gained from the data.

\subsection{Algorithms for Robotic Tasks}
 
The key characteristics of each learning algorithm, their computational intricacies and applications in different robotic tasks in the literature are summarised in Table \ref{tabu}. For the time being, DMP and BC are the most commonly used approaches in MP and EA, respectively. DMP's popularity in robotics lies in its formulation and ease of implementation from the control point of view. Its formulation considers the second-order derivatives, which enable the exploitation of the dynamic behaviour of robots at the torque level. Moreover, DMP's computational complexity linearly rises with the number of task descriptors $(n)$, making it relatively low compared to its peers. Thus, DMP is very easy to implement. For many complex robotic tasks, BC is the most used member of its own class, which initializes either the policy or value function from demonstrations. Its popularity is mainly due to its simplicity and local task exposure. However, the priors-over-skills (PoS) strategy is the most effective, enclosing both the actions and their effects on the learning process. In contrast to Transfer Learning, it also preserves fundamental interaction properties to avoid high bias and variance problems. While encoding the targeted task, Transfer Learning performance drastically subverts for the primary tasks. Besides, working correctly in all changing task conditions requires an additional domain adaptation and task embedding module.  

\subsection{Experimental Benchmarking}

In the framework of LfD, the underlined methods were applied to reconstruct a standard peg-in-hole task (Fig. \ref{demonstrator}), which gives an insight into their performance. Withstanding its simplicity, the peg-in-hole involves all the substantial exchanges between robot and object, including grasping, exploration, and manipulation. The statistical outcomes for all indexes are presented in Table. \ref{task_quality}.

In this case study, an Intel i5 6500U @ 2.3 GHz system with 16 GB RAM and NVIDIA GTX 1650 Ti 4GB was utilized for training and testing the algorithms. Six different human demonstrations with varying initial conditions were collected$\footnote{The dataset is available at \url{https://drive.google.com/drive/u/2/folders/1Va4EBecdNG7zFUtaUqKxuBC4IaRdvI3f}}$. MP algorithms were applied using their predefined settings from the peers. The Proximal Policy Optimization (PPO) technique was used for EA. PPO was chosen for multiple reasons, including faster convergence, better sampling efficiency, and ease of tuning. Henceforward, PPO has been widely applied to different manipulation tasks \cite{Rajeswaran2017,GarciaHernando2020,Yoneda2021,Arunachalam2022}.

As shown in Table \ref{task_quality}, out of MP methods, DMP relatively outperforms the others as DMP ensures not only the adaptive behaviour for varying task conditions, but also inherits key characteristics from ergodic control for better exploration. This task starts with grasping a peg from the top, bringing it all the way to the hole by following the desired path. Finally, explore the hole with the bottom of the peg to align and then insert it. Although, the results achieved by VMP are also promising but since the given task does not involve explicit superposition and modulation of conflicting trajectories for intermediary and final goal points so it eventually under-performs the DMP. 

In case of EA, the trained PPO agent tries to take correlated consecutive steps to remain within the trusted region. Although, being robust against environmental noise and task uncertainties during exploration minimizes the entropy but it requires frequent tuning of hyper-parameters and if not properly configured, suffers from instability in reaching the task goal (i.e, properly positioning the peg into the hole). It was also observed that during certain trajectory rollouts, the PPO misses the target (hole) and left the object (peg) on sideways, which was primarily due to its poor policy initialization.    
    
It must be noted that the smaller the values of $rMSE$ and $IL$ indices, the better the task reproduction of the tested algorithm becomes. Conversely, for  $R^2$, it is preferable to have a higher value to regenerate it closer to the task's base solution (kinaesthetic trajectory). Although none of the single approaches can give the best results for all three metrics, the most promising ones are from DMP for movement primitives and Behavioural Cloning and Priors over Skills from experience abstraction.

\begin{table*}
\label{tabu}
\caption{Summary of discussed techniques in terms of their key characteristics and computational complexity.}
 \centering
\begin{tabular}{|m{4em}|m{4cm}|m{3.0cm}|m{3.5cm}|m{1.6cm}|m{1.65cm}|m{0.5cm}}

\hline
\textbf{Method}&\multicolumn{5}{|c|}{\textbf{Attributes}}  \\
\cline{2-6} 
	 \centering \textbf{list} & \centering \small \textbf{\textit{Key feature}}&\centering \small \textbf{\textit{Pros}}&\centering \small \textbf{\textit{Cons}} &\small \textbf{\textit{Complexity}} &\small \textbf{\textit{Applications}} \\
 \hline 
 \multicolumn{6}{|c|}{\textbf{Movement Primitives}} \\
 \hline
\small DMP & \scriptsize Trains by a demonstration; it constructs a task in a series of non-linear canonical dynamical systems in the form of a damper-spring system. & \begin{itemize}[nosep, wide=0pt, leftmargin=*, after=\strut] \scriptsize \item Time independent;
\scriptsize \item Scaling flexibility;
\scriptsize \item Stability;
\scriptsize \item Needs only a single demonstration.
\end{itemize} & \begin{itemize}[nosep, wide=0pt, leftmargin=*, after=\strut]\vspace{1mm}
\scriptsize \item Lack of Via-point modulation possibility;
\scriptsize \item It is not multitasking;
\scriptsize \item Discontinue inter-primitive transition;
\scriptsize \item Missing robust performance in perturbation;
\scriptsize \item Takes only mono-dimensional temporal inputs .
\end{itemize} &\scriptsize $\mathcal{O}(n)$ & \begin{itemize}[nosep, wide=0pt, leftmargin=*, after=\strut]
\scriptsize \item Interactive rehabilitation\cite{Ijspeert2001};
\scriptsize \item Contact-aware Locomotion\cite{Nakanishi2004};
\scriptsize \item Prehensile Motions\cite{Colome2018}\cite{Zhao2018};
\scriptsize \item Context-driven Path traversing\cite{Kormushev2011a}\cite{Ude2010};
\end{itemize} \\
\hline 
\small PMP \vspace{+0mm}& \scriptsize It Encodes a task by parameterizing it locally using a probabilistic model that is built incrementally based on a mixture of Gaussian distributions or Hidden Markov states.    & \begin{itemize}[nosep, wide=0pt, leftmargin=*, after=\strut] \scriptsize \item Input can be temporal or spatial;
\scriptsize \item Controlled transition and duration to each state;
\scriptsize \item Better task convergence;
\end{itemize} & \begin{itemize}[nosep, wide=0pt, leftmargin=*, after=\strut]\vspace{1mm}
\scriptsize \item Discontinuity on intermediate points;
\scriptsize \item Requires more demonstrations;
\scriptsize \item More hyper-parameters to tune.
\end{itemize} & $\mathcal{O}(n^2)$ & \begin{itemize}[nosep, wide=0pt, leftmargin=*, after=\strut]
\scriptsize \item Household manipulations\cite{Kim2017}\cite{Bozcuoglu2019};
\scriptsize \item  human-robot collaboration\cite{Rozo2014}\cite{Rozo2015};
\scriptsize \item Guesture recognition\cite{Kwon2008}\cite{Lee2010};
\scriptsize \item Stable walking\cite{Rozo2011}, adaptive manipulations\cite{Tanwani2016};
\end{itemize} \\
\hline
\small ProMP \vspace{-2mm}& \scriptsize It takes advantage of the modular architecture of DMP whilst increasing its functionality by capturing each primitive using a single Gaussian at a time. ProMP keeps doing this until it modulates the final Via-point. This cultivates into a stochastic optimal feedback controller.   & \begin{itemize}[nosep, wide=0pt, leftmargin=*, after=\strut] \scriptsize \item Multitasking;
\scriptsize \item Smooth inter-primitive transition;
\scriptsize \item Robust towards uncertainties;
\scriptsize \item Spatiotemporal synchronization.
\end{itemize} & \begin{itemize}[nosep, wide=0pt, leftmargin=*, after=\strut]\vspace{1mm}
\scriptsize \item No exterapolation possibilities;
\scriptsize \item Multiple nodes can not be abstracted at the same time ;
\scriptsize \item Numerical issues if the input features vector is not proportional to the observation size;
\scriptsize \item It requires an inclusive dataset for a proper via-point adaptation;
\end{itemize} & $\mathcal{O}(log(n))$ & \begin{itemize}[nosep, wide=0pt, leftmargin=*, after=\strut]
\scriptsize \item Co-manipulation of objects and assembly\cite{Maeda2017};
\scriptsize \item Playing stroke games\cite{Paraschos2013}\cite{Rueckert2015};
\end{itemize} \\
\hline
\small VMP \vspace{+0mm}& \scriptsize For a given learnable set of wights $mathcal{w}$, it evaluates if the given Via-point falls within a manually-chosen threshold $\eta$ $p(y_{via}|\mu_{\omega},\Sigma_{\omega})>\eta$ . Consequently, it uses the modulation function $f(x)$ to modulate it. Otherwise, the elementary trajectory function $h(x)$ brings the whole trajectory distribution to the new target ($y(x)=h(x)+f(x)$). & \begin{itemize}[nosep, wide=0pt, leftmargin=*, after=\strut] \scriptsize \item Extrapolation capacity is included;
\scriptsize \item A straightforward implementation that takes an MP and a few via-point to capture the task;
\scriptsize \item Better accuracy;
\end{itemize} & \begin{itemize}[nosep, wide=0pt, leftmargin=*, after=\strut]\vspace{0mm}
\scriptsize \item Obstacle avoidance is not as expected.
\scriptsize \item Controller stability is compromised.
\scriptsize \item Local spare representation makes it prone to noise.
\end{itemize} & $\mathcal{O}(nlog(n))$ & \begin{itemize}[nosep, wide=0pt, leftmargin=*, after=\strut]
\scriptsize \item Human-centric grasping, path traversing \cite{Zhou2019};
\end{itemize} \\
\hline 
\small FMP & \scriptsize GP conduct regression by defining a distribution over all possible functions that fit the dataset. GP uses basis functions to apply linear regression for wiggly datasets that disturbs the balance of the size-feature ratio of the input and uses kernel trick to deal with the issue. & \begin{itemize}[nosep, wide=0pt, leftmargin=*, after=\strut] \scriptsize 

\scriptsize \item More accurate regression;
\scriptsize \item Simple math makes it easy to implement.
\scriptsize \item Better fit without cross validation.

\end{itemize} & \begin{itemize}[nosep, wide=0pt, leftmargin=*, after=\strut]\vspace{0mm}
\scriptsize \item Usually it is computationally expensive.
\scriptsize \item Requires more information for prediction
\scriptsize \item Compromises efficiency in high dimensional space.
\end{itemize} &  $\mathcal{O}(n^3)$ &  \begin{itemize}[nosep, wide=0pt, leftmargin=*, after=\strut]
\scriptsize \item Trajectory tracking and reaching\cite{Forte2012};
\scriptsize \item bipedal Locomotion\cite{Werner2015};
\scriptsize \item robust environment interaction\cite{Schreiter2015};
\end{itemize} \\
\hline
\small KMP & \scriptsize KMP is a kernel-based approach that separately solves the optimization problem for $\Sigma$ and $\mu$, reducing the difference between a parametric trajectory and the reference one. & \begin{itemize}[nosep, wide=0pt, leftmargin=*, after=\strut] \scriptsize 

\scriptsize \item Better extrapolation possibilities;
\scriptsize \item Requires lesser demonstration, specially compared with ProMP;
\scriptsize \item Deals with dimensionality issues of input with high number of features.
\end{itemize} & \begin{itemize}[nosep, wide=0pt, leftmargin=*, after=\strut]\vspace{1mm}
\scriptsize \item Usually it is computationally expensive;
\scriptsize \item The accuracy of the approach highly bounds with the type of kernel;
\scriptsize \item In the absence of proper prior information on the target position, trajectory adaptation will face problems.
\end{itemize} &  $\mathcal{O}(n^3)$ &  \begin{itemize}[nosep, wide=0pt, leftmargin=*, after=\strut]
\scriptsize \item Object transportation, Human-robot collaboration\cite{Huang2019}\cite{Silverio2019};
\scriptsize \item Dexterous Manipulation\cite{Katyara2021}\cite{Katyara2021a};
\scriptsize \item Non-holonomic motion planning\cite{Deng2021};
\end{itemize} \\
\hline
\end{tabular}
\end{table*}
\begin{table*}
\label{tabu}
 \centering
\begin{tabular}{|m{4em}|m{4cm}|m{3.0cm}|m{3.5cm}|m{1.6cm}|m{1.6cm}|m{0.5cm}}

\hline
\textbf{Method}&\multicolumn{5}{|c|}{\textbf{Attributes}}  \\
\cline{2-6} 
	 \centering \textbf{list} & \centering \small \textbf{\textit{Key feature}}&\centering \small \textbf{\textit{Pros}}&\centering \small \textbf{\textit{Cons}} &\small \textbf{\textit{Complexity}} &\small \textbf{\textit{Applications}} \\
\hline 
\multicolumn{6}{|c|}{\textbf{Experience Abstraction}} \\
\hline 
\small Skill Transfer & \scriptsize Shares learned task features with other approaches to not only reduce search space but also ensures better adaptation. & \begin{itemize}[nosep, wide=0pt, leftmargin=*, after=\strut] \scriptsize 

\scriptsize \item Alleviate the need for extensive training;
\scriptsize \item Suitable for dealing with exploration-exploitation dilemma;
\scriptsize \item Less prone to task over-fitting;

\end{itemize} & \begin{itemize}[nosep, wide=0pt, leftmargin=*, after=\strut]\vspace{0mm}
\scriptsize \item Sparse representation affects task accuracy.
\scriptsize \item Kinematic-dynamic constraints on mapping between agents.
\scriptsize \item Applicable only to tasks having similar initial and finial conditions.
\end{itemize} &  $\mathcal{O}(n)$ &  \begin{itemize}[nosep, wide=0pt, leftmargin=*, after=\strut]
\scriptsize \item Object location\cite{Karaoguz2019}\cite{Weng2020};
\scriptsize \item In-hand manipulation\cite{Bocsi2013}\cite{Lee2021};
\scriptsize \item Dual-arm dexterous manipulations\cite{Seita2019}\cite{Schwarz2018};
\end{itemize}  \\

\hline
\small Behavioural Cloning & \scriptsize Helps to initialize the policy using human demonstrations in off-line manner for improvemed task convergence and performance. & \begin{itemize}[nosep, wide=0pt, leftmargin=*, after=\strut] \scriptsize 

\scriptsize \item Incorporate human intuition in executing a task;
\scriptsize \item Leads to achieve optimal control;
\scriptsize \item Simple but effective strategy for direct policy learning;

\end{itemize} & \begin{itemize}[nosep, wide=0pt, leftmargin=*, after=\strut]\vspace{0mm}
\scriptsize \item Data need to be independent and identically distributed.
\scriptsize \item Treats decision making process as prediction problem.
\scriptsize \item Less generalizable to task with varying dynamics.
\end{itemize} &  $\mathcal{O}(nlog(n))$ &  \begin{itemize}[nosep, wide=0pt, leftmargin=*, after=\strut]
\scriptsize \item Furniture assembly\cite{Niekum2015};
\scriptsize \item Environment exploration\cite{Kadous2005};
\scriptsize \item Autonomous locomotion\cite{Desai2020}.
\end{itemize}  \\

\hline
\small Priors Over Skills & \scriptsize Extract generalized task-related features from the human data to build a library of skills for exhaustive execution of family of tasks. & \begin{itemize}[nosep, wide=0pt, leftmargin=*, after=\strut] \scriptsize 

\scriptsize \item Distributed nature makes it easier to adapt.
\scriptsize \item Generalized behaviour for global stability.
\scriptsize \item Similar performance for both policy and value based agents.

\end{itemize} & \begin{itemize}[nosep, wide=0pt, leftmargin=*, after=\strut]\vspace{0mm}
\scriptsize \item Need for latent representation.
\scriptsize \item Requires more resources.
\scriptsize \item Necessitates sdditional higher dimensional skill embedding.
\end{itemize} &  $\mathcal{O}(n^2)$ &  \begin{itemize}[nosep, wide=0pt, leftmargin=*, after=\strut]
\scriptsize \item Object grasping and stacking\cite{Hausman2018};
\scriptsize \item Navigation, precision grasping, fine manipulation \cite{Pertsch2020};
\end{itemize}  \\

\hline
\end{tabular}
\end{table*}

\begin{table}[t]
\caption{Experimental evaluation of different movement primitives and experience abstraction algorithms using statistical measures.}
\label{task_quality}
\begin{center}
\begin{tabular}{c|c|c|c}
\hline
\textbf{Method} & \textbf{$rMSE$} & \textbf{$R^2$} & \textbf{$IL$} \\
\hline
\textit{DMP} & {0.13876} & {0.26625} & {\textbf{1.02854}} \\
\hline
\textit{PMP} & {0.30452} & {0.40449} & {1.13224} \\
\hline
\textit{ProMP} & {0.27532} & {0.20976} & {1.84252} \\
\hline
\textit{VMP} & {0.12043} & {0.27352} & {1.21465} \\
\hline
\textit{FMP} & {0.22132} & {0.24742} & {1.57432} \\
\hline
\textit{KMP} & {0.19751} & {0.25314} & {1.34211} \\
\hline
\textit{PPO-ST} & {0.07632} & {0.39015} & {1.18764}
\\
\hline
\textit{PPO-BC} & {\textbf{0.02324}} & {0.41783} & {1.09852}
\\
\hline
\textit{PPO-PoS} & {0.05431} & {\textbf{0.42132}} & {1.18653}
\\
\hline
\end{tabular}
\end{center}

\end{table}

\section{Conclusions and Recommendations}

In this review article, we summarised the key traits, strengths, limitations, computational complexity and applications of different robot learning algorithms ranging from motion primitives to skill embedding. We addressed both soft and rigid boundaries of various agents under distinct circumstances that affect their learning performance in realizing specific and generalized motions, skills, and actions. A standard peg-in-hole task was reconstructed and evaluated using different statistical metrics to benchmark and understand the performance of discussed approaches. Based on what we have found, kernel-based and policy-gradient-based approaches for LfD and Incremental Learning outperformed the others in the same order. However, they suffer from computational loads when applied to more complex tasks.  

Because of the limited applications and hidden potential of algorithms for task decomposition, trajectory encoding, motion representation, shared behaviours, task semantic masking, and skill embedding discussed in this paper, the following recommendations are made for potential future directions:

\begin{itemize}
    \item \textbf{Inherit Learning:} Like humans, learning algorithms can pass many of their learnt components and behaviours to others, especially those control systems with common characteristics. Thus, the transferred quasi-equivalent learnt model parameters should adapt to the new ones by some fine tune of their hyperparameter. In particular, it will be beneficial for benchmarking (similar to Q-learning algorithms) \cite{Zhang2021}.
    
    \item \textbf{Attention Modelling:} Developing a generic library of skills from human demonstrations is a necessary but not sufficient
    condition for global task execution. Hence, inferring an attention model from their observations is highly recommended. Attention models will help to map humans' perceptual modality when it gets activated at different task instances. This will help to understand, integrate and exploit multi-sensory information into the robot system for the anthropomorphic execution of tasks, similar to sequence transformers \cite{Vaswani2017}.
    
    \item \textbf{Transceive Learning:} Transferring knowledge from a trained model to its naive counterpart for better exploitation and less exploration is promising to curtail the computation burden and improve performance. However, in specific scenarios, like sim-to-real skill transfer, where there is not enough correspondence between their workspaces, this approach may fall into deficiencies. Instead of randomly initializing the domain parameters in this condition, it is highly recommended to use real-time environment data using different perceptual feedback and train the naive agent (digital twin). The trained policy will guarantee better domain adaptation and generalize well to different task extensions, similar to the skill transfer in real-sim-real adaptation (digital twin) \cite{Liu2022};
    
    \item \textbf{Conditional Bootstrapping:} The problem is addressed sequentially in curriculum learning by exploiting the predicate prior information on primitive actions. However, such predicate prior information induces biases in selecting the following actions (posteriors), which hinders them from being scaled up to complex environments. The conditional bootstrapping instead will phase estimation (i.e., predicting next moves) from the current observation of state (i.e., following Markov property) by conditioning it over the previous moves (i.e., making intuition of what actions have made an agent end up being in this state), i.e., similar to AlphaZero \cite{Zhang2020};
    \item \textbf{Compact Encoding:} Having uniform representation for both continuous and discrete spaces and defining a model to deal with all varying dynamics of the environment for different tasks is similar to what the human brain does. The compact encoding will map linear states to non-linear actions to update its internal parameters at run-time. It will consider task requirements without significantly changing its internal architecture, similar to GATO \cite{Reed2022}.

\end{itemize}

\bibliographystyle{IEEEtran}
\bibliography{TCDS} 

\end{document}